\documentclass[conference]{IEEEtran}
\IEEEoverridecommandlockouts
\usepackage{cite}
\usepackage{amsmath,amssymb,amsfonts}
\usepackage{algorithmic}
\usepackage{graphicx}
\usepackage{textcomp}
\usepackage{xcolor}
\usepackage{siunitx}
\usepackage{subcaption}
\usepackage{booktabs}
\usepackage{todonotes}
\usepackage{xcolor}

\def\BibTeX{{\rm B\kern-.05em{\sc i\kern-.025em b}\kern-.08em
    T\kern-.1667em\lower.7ex\hbox{E}\kern-.125emX}}

\begin{document}

\title{Multi-LiDAR Localization and Mapping Pipeline \\ for Urban Autonomous Driving \\

\thanks{The research was partially funded by the Federal Ministry of Education and Research of Germany (BMBF) within the project Wies'n Shuttle (FKZ 03ZU1105AA) in the MCube cluster, the Bavarian Research Foundation (BFS), and through basic research funds from the Institute of Automotive Technology.}
}

\author{\IEEEauthorblockN{Florian Sauerbeck$^{1}$, Dominik Kulmer$^{1}$, Markus Pielmeier$^{1}$, Maximilian Leitenstern$^{1}$,\\ Christoph Weiß$^{1}$, and Johannes Betz$^{2}$}
\IEEEauthorblockA{
$^{1}$\textit{Institute of Automotive Technology,}
\textit{Munich Institute of Robotics and Machine Intelligence (MIRMI),}\\
\textit{Technical University of Munich,}
85748 Garching, Germany \\
$^{2}$\textit{Professorship Autonomous Vehicle Systems,}
\textit{Munich Institute of Robotics and Machine Intelligence (MIRMI),}\\
\textit{Technical University of Munich,}
85748 Garching, Germany \\
{florian.sauerbeck@tum.de}
}
}

\newcommand\copyrighttext{%
  \footnotesize 978-1-5386-5541-2/18/\$31.00~\copyright2018 IEEE
}
\newcommand\copyrightnotice{%
\begin{tikzpicture}[remember picture,overlay]
\node[anchor=south,yshift=25pt,xshift=-186pt] at (current page.south) 
    {\copyrighttext};
\end{tikzpicture}%
}

\maketitle
\copyrightnotice
\begin{abstract}
Autonomous vehicles require accurate and robust localization and mapping algorithms to navigate safely and reliably in urban environments.
We present a novel sensor fusion-based pipeline for offline mapping and online localization based on LiDAR sensors.
The proposed approach leverages four LiDAR sensors.
Mapping and localization algorithms are based on the \textit{KISS-ICP}, enabling real-time performance and high accuracy.
We introduce an approach to generate semantic maps for driving tasks such as path planning.
The presented pipeline is integrated into the \textit{ROS 2} based \textit{Autoware} software stack, providing a robust and flexible environment for autonomous driving applications.
We show that our pipeline outperforms state-of-the-art approaches for a given research vehicle and real-world autonomous driving application.
\end{abstract}

\begin{IEEEkeywords}
Autonomous Vehicles, LiDAR, Maps, Sensor Fusion, SLAM
\end{IEEEkeywords}

\section{Introduction}

Robust and accurate localization plays a central role in the realm of urban autonomous driving \cite{pendleton2017perception}. 
Especially in urban areas with high surrounding buildings, it is necessary to have a localization system that does not rely solely on the Global Navigation Satellite System (GNSS).
This is where the Simultaneous Localization And Mapping (SLAM) problem comes into play. The go-to sensor for addressing the SLAM task is the Light Detection And Ranging (LiDAR) sensor \cite{kuutti2018survey}, which is widely used in the context of autonomous driving.
Planning a trajectory and driving through these mapped environments requires additional semantic maps, including lane information.

\subsection{Related Work}

\subsubsection{LiDAR SLAM}
LiDAR SLAM approaches showed convincing results with different approaches throughout the last years.
Some of the best-performing algorithms are \textit{LOAM} \cite{zhang2014loam}, \textit{MULLS} \cite{pan2021mulls}, and \textit{CT-ICP} \cite{dellenbach2022ct}.
The \textit{KISS-ICP} algorithm \cite{vizzo2023} is an Iterative Closest Point (ICP) approach with a focus on easy adaptability.
It promises to work well out of the box for several applications and use cases and was developed for real-time and real-world applications.
However, \textit{KISS-ICP} is a LiDAR odometry algorithm and thus does not offer loop closure or backend optimization.
To add such constraints, the \textit{Interactive SLAM} \cite{koide2020interactive} offers a toolbox to manually add them and run additional registration and optimization.

\begin{figure}[t!]
    \centering
    \includegraphics[angle=-0, width=0.7\linewidth, trim={0cm 4.5cm 0cm 3.25cm}, clip]{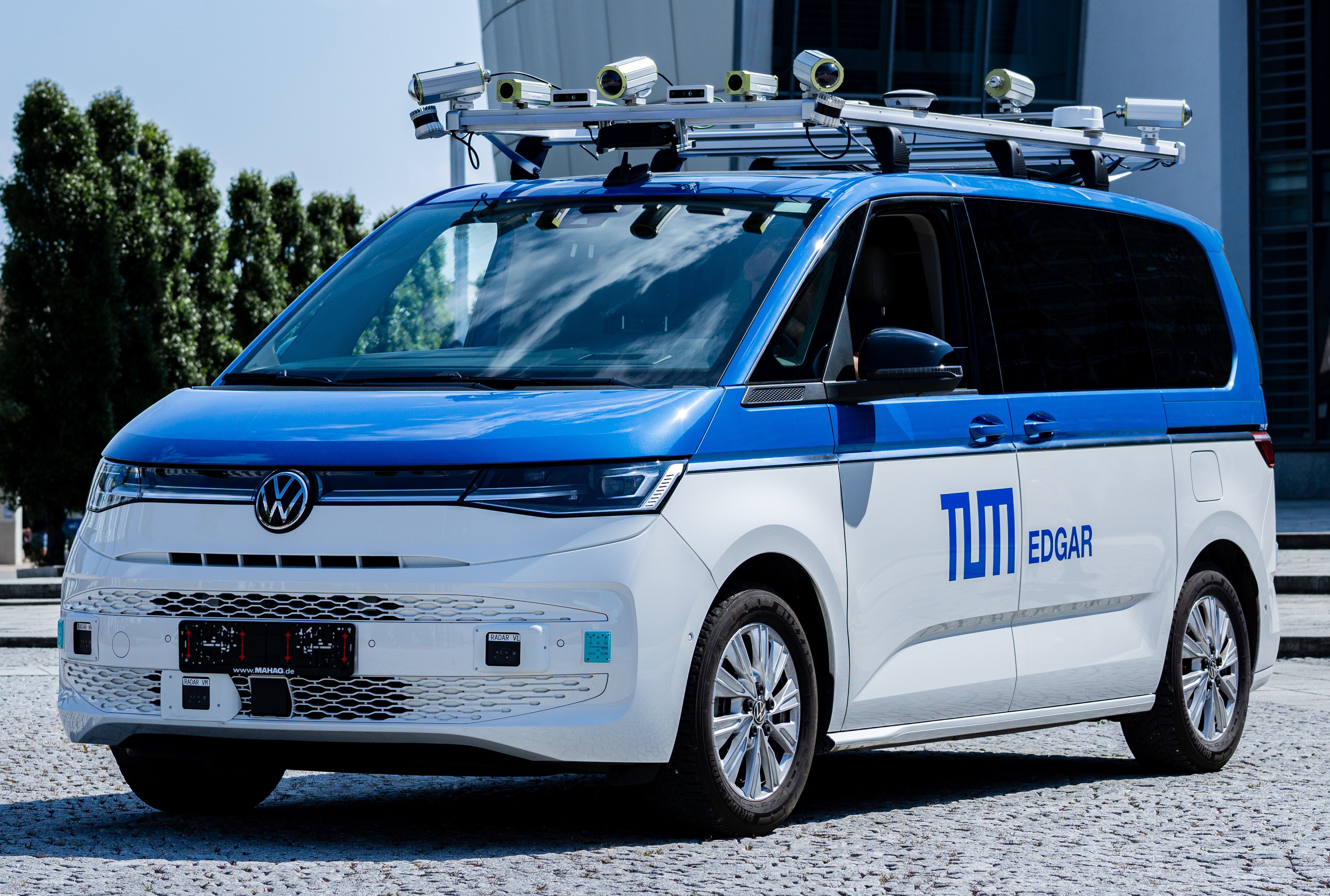}
    \caption{Research vehicle EDGAR.}
    \label{fig:edgar}
\end{figure}

\subsubsection{Multi-LiDAR Localization}
Some works have already dealt with multi-LiDAR systems and their additional challenges.
The main difference from single-LiDAR is that spatial calibration and temporal synchronization of sensor clocks are crucial.
\textit{M-LOAM} \cite{jiao2021mloam} is an extension of \textit{LOAM} \cite{zhang2014loam}. It adds multi-LiDAR support and also the ability for online calibration.
Other works focused on LiDAR-LiDAR calibration, which is essential for fusing point clouds from different sensors \cite{Yan.2022, jiao2019novel, gao2010line}.

\subsubsection{Mapping for Autonomous Vehicles (AVs)}
In addition to 3D point cloud maps, most autonomous driving software stacks are based on High Definition (HD) maps that provide semantic information \cite{seif2016autonomous}.
A huge challenge is to generate these maps at scale and keep them up-to-date \cite{kim2021hd}.
A review of automated methods utilizing onboard sensor data and aerial images can be found in \cite{bao2022high}.
Semantic mapping approaches are either deep learning-based \cite{li2022hdmapnet, liu2022vectormapnet} or based on conventional algorithms \cite{zhang2018road, sun20193d}.
\textit{Lanelet2} \cite{poggenhans2018lanelet2} is a widely used lane-level map format.

\subsection{Vehicle Setup}
\begin{figure}[ht!]
    \centering
    \includegraphics[angle=-0, width=0.7\linewidth, trim={0cm 0cm 0cm 0cm}, clip]{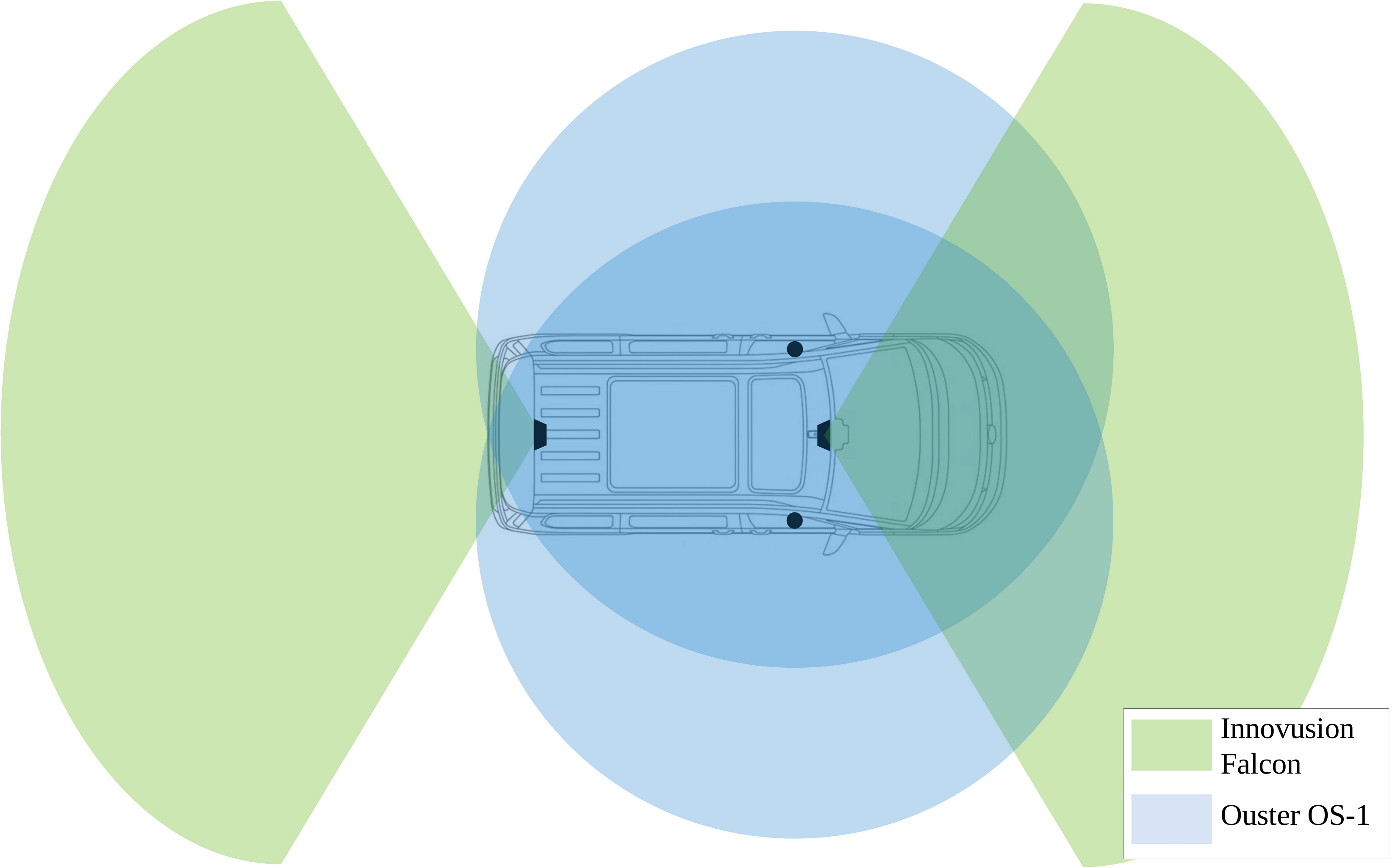}
    \caption{EDGAR LiDARs' FOV in bird's eye view.}
    \label{fig:fov}
\end{figure}

The autonomous research vehicle EDGAR (Fig. \ref{fig:edgar}) is used to realize this work.
It has a total of four LiDARs: two \textit{Innovusion Falcon Kinetic} for long-range to the front and rear, and two \textit{Ouster OS1-128} tilted by \SI{20}{\degree} on the sides to cover the close and mid-range around the vehicle. Fig. \ref{fig:fov} shows the LiDARs' field of view (FOV).
The comparably high amount of LiDARs was chosen as part of the overall vehicle concept to drive safely through urban and crowded areas with many close objects, e.g. pedestrians.
Thus, the localization module must fuse the LiDARs for reliable and robust localization.
For reference localization in areas with GNSS coverage, an RTK-corrected differential GNSS system can be used.
Table \ref{tab:hw} lists the relevant hardware components.

\begin{table}
\centering
\scriptsize
\caption{Overview of the perception hardware components.}
\label{tab:hw}
    \begin{tabular}{c|c|c|c}
    \toprule
        \textbf{Component} & \textbf{Manufacturer} & \textbf{Model} & \textbf{FOV (h x v)} \\
        \midrule
        LiDAR Mid-Range & \textit{Ouster} & \textit{OS1-128} & \SI{360}{\degree} x \SI{45}{\degree} \\
        LiDAR Long-Range & \textit{Innovusion} & \textit{Falcon Kinetic} & \SI{120}{\degree} x \SI{25}{\degree} \\
        GNSS-IMU & \textit{Novatel} & \textit{PwrPak7D-E2} & - \\
    \bottomrule
    \end{tabular}
\end{table}

\section{Methodology}
\begin{figure}[ht!]
    \centering
    \includegraphics[width=0.90\linewidth, trim={0cm 0cm 0cm 0cm}, clip]{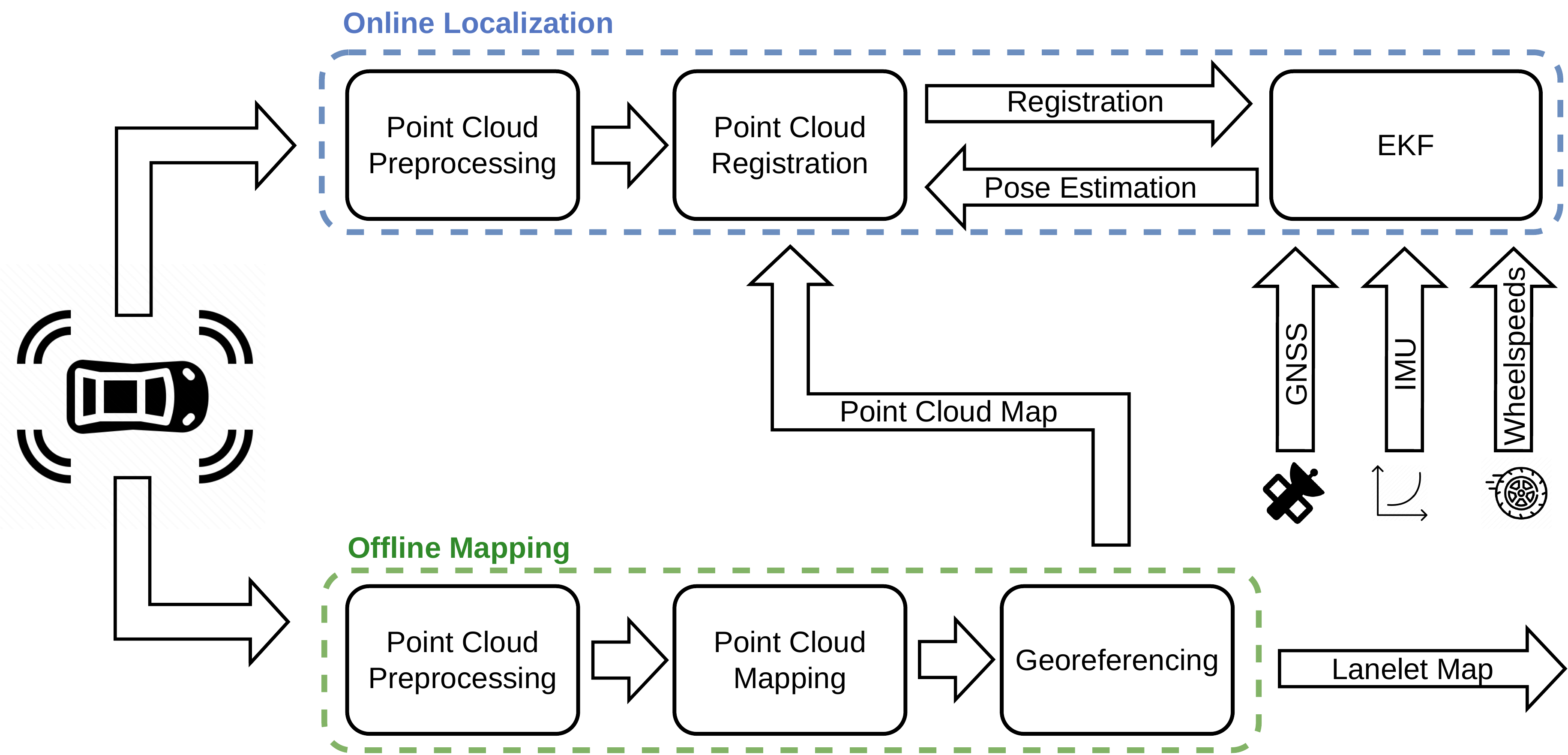}
    \caption{Overview of the presented multi-LiDAR localization and mapping pipeline.}
    \label{fig:pipeline}
\end{figure}

Our overall pipeline is shown in Fig. \ref{fig:pipeline}.
After preprocessing the raw data, a 3D point cloud map and a semantic map are built and referenced to the GNSS trajectory offline.
The point cloud map is used online to localize the vehicle by scan registration.
An Extended Kalman Filter (EKF) fuses all information and outputs high-frequency ego-states.
The pipeline is integrated into our custom \textit{ROS 2} environment based on the \textit{Autoware}\footnote{https://www.autoware.org/} software stack.

\subsection{Sensor Fusion}
The sensor fusion strategy is crucial to the performance of the pipeline.
Before processing the point clouds, preprocessing steps, e.g. voxel filtering, are applied to reduce the amount of data.
Also, deskewing has to be applied to reduce the motion blur of the LiDAR scans.
Many problems stem from inaccuracies in the sensor data's spatial calibration or temporal synchronization.
To minimize these problems, we run separate registrations for each LiDAR scan rather than fusing all point clouds before registration.
Each scan is first transformed into a joint frame and then matched with the map.
In this way, we do not have to take care of temporal differences between the single scans, and miscalibrations can also be easily compensated for.

\subsection{Mapping}
The mapping pipeline consists of a 3D point cloud mapping and a semantic mapping approach to extract lane-level information.

\subsubsection{Point Cloud Mapping}
To select the mapping method, the previously mentioned SLAM algorithms were evaluated. Due to the atypical LiDAR setup and the above-mentioned associated challenges, state-of-the-art LiDAR SLAM algorithms failed to create consistent maps.
\textit{KISS-ICP} with separated scan registration showed robust behavior with respect to map creation.
The map is post-processed with the \textit{Interactive SLAM} \cite{koide2020interactive}. This framework allows performing loop closure and graph optimization based on individual poses along the trajectory.

\subsubsection{Semantic mapping}
\begin{figure}[t!]
    \centering
    \includegraphics[width=0.90\linewidth, trim={0cm 0cm 0cm 0cm}, clip]{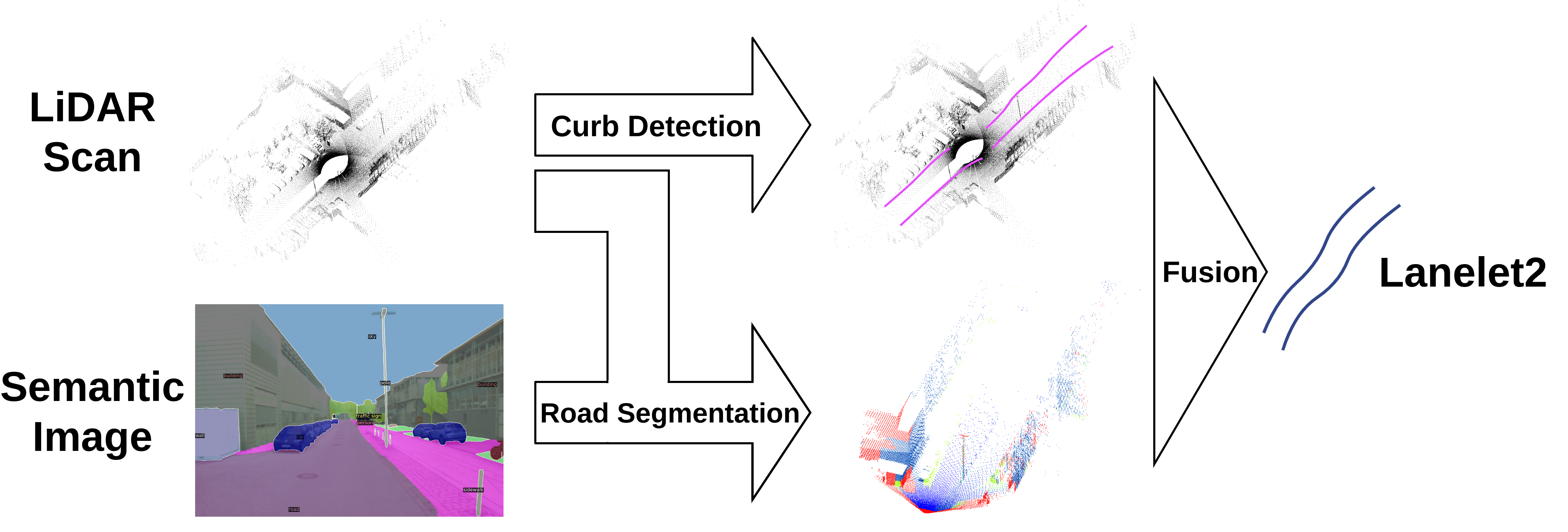}
    \caption{Extraction of semantic street map from LiDAR point cloud and semantic image.}
    \label{fig:lane_extraction}
\end{figure}

Semantic maps are needed for path and behavior planning.
The aim of the semantic mapping pipeline is to run as automated as possible.
However, a manual semantic mapping pipeline is still needed to correct wrong-marked parts and generate ground-truth data.
For \textbf{manual semantic mapping}, the \textit{Vector Map Builder}\footnote{https://tools.tier4.jp/feature/vector\_map\_builder\_ll2/} is used.
Point cloud maps are annotated manually by marking lane boundaries, etc.
An \textbf{automated semantic mapping} pipeline is being developed for better scalability. The point cloud map undergoes intensity-based filtering to extract white road markings. Curbs are detected using a sliding-window approach along the driven trajectory. These two essential road boundary elements are fused in a prior semantic map.
In the absence of lines and curbs, semantic information is used to extract the road boundary.
Therefore, the LiDAR scans are painted with semantic labels by projecting them into a segmented 2D camera image before building the point cloud map.
State-of-the-art 3D LiDAR segmentation algorithms did not work sufficiently well due to the multi-LiDAR setup.
The borders of the areas segmented as roads are then fused with the existing semantic lane map.
The automated semantic mapping process is visualized in Fig. \ref{fig:lane_extraction}.

\subsubsection{Georeferencing}
Since no GNSS is used during the mapping process, the map must be georeferenced to allow GNSS use during localization.
After a UTM projection \cite{utmproj} to local cartesian coordinates, the trajectories are roughly aligned using the \textit{Umeyama} algorithm \cite{umeyama1991least}.
A predefined number of corresponding control points from each trajectory is manually selected. Based on these points, a piecewise linear rubber-sheet transformation \cite{MarvinS.WhiteJr.1985} is used to match the trajectories. The transformations are then applied to the map. A projection to global coordinates provides a georeferenced map. This also enables automatic enrichment and evaluation of the semantic map using information from \textit{OpenStreetMap}\footnote{https://www.openstreetmap.org/}. Note that the rubber-sheet transformation only works in 2D. Therefore, the height information is currently not adjusted.

\subsection{Online Localization}
Following the mapping process results, the localization was also based on the \textit{KISS-ICP}. Therefore, the algorithm has been extended.
In localization mode, the local map used in the standard implementation is replaced by the a priori-created point cloud map to allow for reliable localization without drift.
Furthermore, the algorithm can use estimates from an external state estimator for the initial pose estimation of the point cloud registration.
An Extended Kalman Filter (EKF) was implemented based on \textit{Autoware} to fuse LiDAR registration, GNSS, IMU, wheel speed information, and vehicle dynamics model assumptions.

\section{Results and Discussion}

\subsection{Results}
\begin{figure}[ht!]
    \centering
    \includegraphics[angle=-0, width=0.95\linewidth, trim={0cm 0cm 0cm 0cm}, clip]{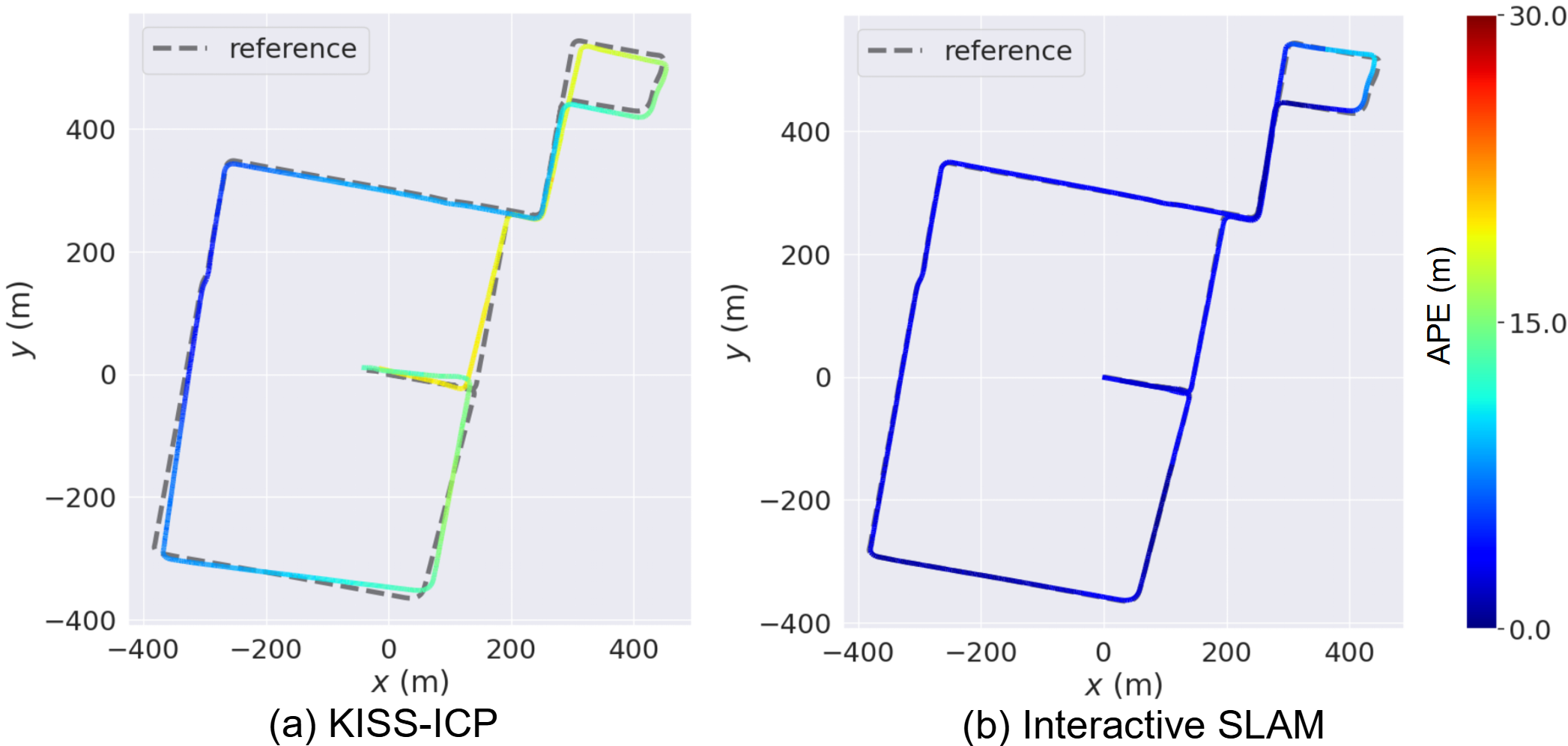}
    \caption{APE of trajectories from \textit{KISS-ICP} (a) and post-processed with \textit{Interactive SLAM} (b). }
    \label{fig:interactive}
\end{figure}

\begin{figure}[ht!]
    \centering
    \includegraphics[angle=-0, width=0.75\linewidth, trim={0cm 2cm 0cm 1cm}, clip]{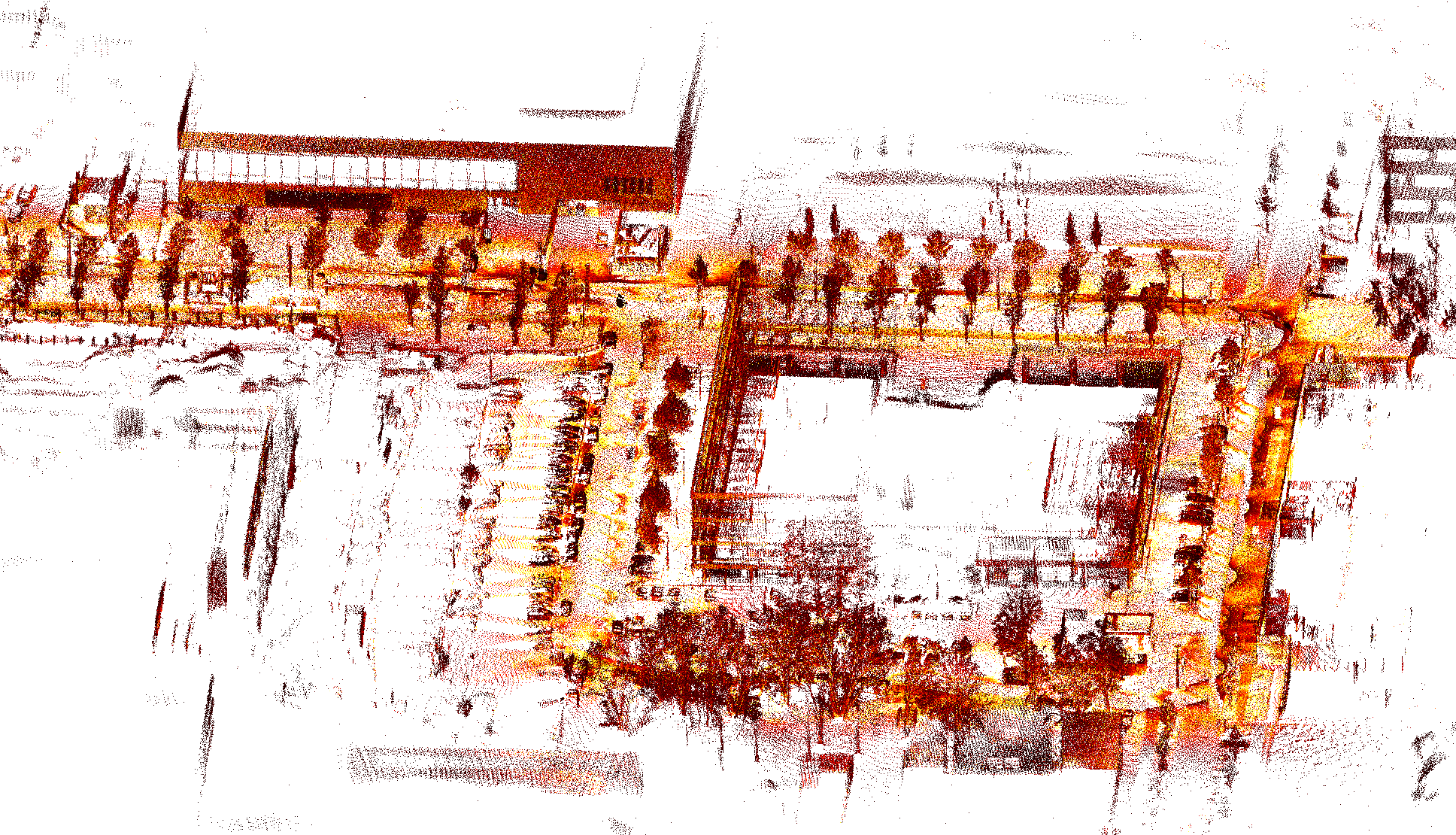}
    \caption{Cutout of the generated point cloud map at the TUM Campus Garching.}
    \label{fig:map}
\end{figure}

The presented pipeline enables the creation of accurate and detailed HD maps for autonomous driving. 
The map created offline by the \textit{KISS-ICP} shows a high consistency between the registered point cloud frames.
However, a significant amount of drift accumulates over the entire length of the trajectory.
Post-processing in the \textit{Interactive SLAM} can significantly minimize the Absolute Position Error (APE). In our example (Fig. \ref{fig:interactive}), the Root Mean Squared Error (RMSE) of the APE was reduced from \SI{12.7}{\meter} to \SI{3.6}{\meter}.
Georeferencing allows the created maps to be referenced globally. This enables the use of external absolute position sources, such as GNSS, in combination with relative estimates from the LiDAR registration for localization.
Fig. \ref{fig:map} shows a point cloud map generated with the presented pipeline.

The automated semantic mapping pipeline currently relies mainly on road markings and curbs. It will have to be enhanced in future work for a better generalizing approach.

\subsection{Discussion}
Despite the good results in creating the maps and the robust approach to localization, our pipeline still has shortcomings.
Single registrations of LiDAR sensors can compensate for errors in synchronization and external calibration.
Still, it results in a noisy trajectory, as poses jump between the frames of badly calibrated sensors.
The additional computational requirements will have to be investigated in more detail.
Our georeferencing relies on a reference trajectory, which requires a reliable GNSS signal. The precision of the result depends directly on the standard deviation of the external position signal.
Furthermore, the performance of semantic information extraction strongly depends on the chosen scenario. Comprehensive quantification of real-world data is currently not possible due to a lack of labeled data.
The pipeline has only been tested in a limited environment around the TUM campus. For a better statement of the robustness of the whole pipeline, an evaluation of various real-world scenarios is necessary.

\section{Conclusion and Future Work}

In this work, we presented a pipeline for multi-LiDAR mapping and localization for AVs.
Our approach to sensor fusion and separated LiDAR registration has proven robust compared to state-of-the-art SLAM algorithms.
Post-processing the point cloud maps allowed further improvement of our results and associating them with a global reference.
We showed how different data sources can be merged to add a semantic layer to the map.
Using multi-modal localization within a static map, reliable localization in urban road traffic, even in GNSS-denied environments is possible.

To improve the robustness and accuracy of the approach, an online multi-LiDAR calibration approach is to be developed.
Also, more comprehensive evaluation is necessary to fully quantify our approach.

\clearpage

\bibliographystyle{IEEEtran}
\bibliography{references}

\end{document}